\documentclass[accepted]{uai2026} 
\usepackage[american]{babel}
\usepackage{natbib}
\usepackage{mathtools} 
\usepackage{booktabs} 
\usepackage{tikz} 
\usepackage{amsmath}
\usepackage{algorithm}
\usepackage{amsmath}
\usepackage{amssymb}
\usepackage{mathtools}
\usepackage{amsthm}
\usepackage{float}
\usepackage{multirow}
\usepackage{algorithmic}
\usepackage{enumitem}
\usepackage{cleveref}

\usepackage[textsize=tiny]{todonotes}

\theoremstyle{plain}
\newtheorem{theorem}{Theorem}[section]
\newtheorem{proposition}[theorem]{Proposition}
\newtheorem{lemma}[theorem]{Lemma}
\newtheorem{corollary}[theorem]{Corollary}
\theoremstyle{definition}
\newtheorem{definition}[theorem]{Definition}
\newtheorem{assumption}[theorem]{Assumption}
\theoremstyle{remark}
\newtheorem{remark}[theorem]{Remark}

\def\RR{\mathbb{R}}
\def\CC{\mathcal{C}}
\def\DC{\mathcal{D}}
\newcommand{\E}{\mathbb{E}}
\newcommand{\KL}{\mathrm{KL}}
\newcommand{\CE}{\mathrm{CE}}

\newcommand{\ip}[2]{\langle #1, #2 \rangle}

\def\Dpub{\mathcal{D}_{\text{pub}}}
\newcommand{\mycomment}[1]{}

\theoremstyle{plain}
\theoremstyle{definition}
\theoremstyle{remark}

\newcommand{\PDi}{\mathcal{P}_i}

\newcommand{\PDpubx}{\mathcal{P}_{\mathrm{pub},x}}

\newcommand{\pstar}{p^*}
\newcommand{\pagg}{p_{\mathrm{agg}}}
\newcommand{\Jagg}{J_{\mathrm{agg}}}
\newcommand{\Jstar}{J_*}

\newcommand{\Dpubx}{\mathcal{D}_{\mathrm{pub},x}}

\title{Who to Trust? Aggregating Client Predictions in Federated Distillation}

\author[1]{Viktor Kovalchuk}
\author[2]{Denis Son}
\author[1]{Arman Bolatov}
\author[1]{Mohsen Guizani}
\author[1]{Samuel Horv\'ath}
\author[1]{Maxim Panov}
\author[1]{Martin Tak\'a\v{c}}
\author[1]{Eduard Gorbunov}
\author[1]{Nikita Kotelevskii}
\affil[1]{%
    Mohamed bin Zayed University of Artificial Intelligence (MBZUAI)\\
    Abu Dhabi, UAE
}
\affil[2]{%
    Independent Researcher
}
  
\begin{document}
\maketitle

\begin{abstract}
Federated Distillation enables distributed learning for clients with heterogeneous model architectures. In this paradigm, the server and clients exchange predictions on a shared unlabeled public dataset, rather than model parameters or gradients.
However, under data heterogeneity (e.g., \emph{class mismatch}), clients produce unreliable predictions for instances from unfamiliar classes.
An equally weighted combination of such predictions corrupts the teacher signal used for distillation.
In this paper, we theoretically analyze Federated Distillation and show that aggregating client predictions on a shared public dataset converges to a neighborhood of the optimum, with the neighborhood size controlled by the aggregation quality.
We propose two uncertainty-aware aggregation methods, \textbf{UWA} and \textbf{sUWA}, that use density-based estimates to down-weight unreliable client predictions.
Experiments on image and text classification datasets confirm that our methods are most effective under high data heterogeneity, while matching standard averaging when heterogeneity is low. Code is available at
\url{https://github.com/kovalchuk026/fd_aggregators}.
\end{abstract}

\section{Introduction}
\label{sec:introduction}

Federated Learning (FL) is a distributed learning paradigm that allows multiple clients to collaboratively train their models without sharing private data~\citep{mcmahan2017fedavg,konevcny2016federated,kairouz2021advances,wang2021field}.
However, most FL algorithms require sharing model parameters or gradients, which can be prohibitively expensive for modern deep networks.

To reduce communication costs, several works explore an alternative paradigm called Federated Distillation (FD)~\citep{li2019fedmd,shao2024selective,wei2022knowledge}.
In FD, clients share model outputs (e.g., logits or predicted probabilities) on a public dataset, rather than model parameters.
This setup allows for a significant reduction in communication and naturally supports heterogeneous client architectures, since FD only requires the output spaces of client models to match.
As a result, the core algorithmic challenge shifts from \emph{parameter aggregation} to \emph{logits or probabilities aggregation}, which are used for distillation.

However, standard FD approaches\citep{li2019fedmd,lin2020ensemble} treat all client predictions as equally reliable, which becomes problematic when the data are heterogeneous.
In this work, we focus on a practically important form of such heterogeneity, a label shift that leads to \emph{class mismatch}. In this setting, each client observes only a subset of the global label space, while the public dataset contains samples from all classes.
Consequently, some public samples belong to classes absent from a given client's training data. For such samples, the client effectively predicts out-of-distribution (OoD) inputs, and the resulting predictions can be misleading.
Naive averaging of those predictions, therefore, may severely degrade the quality of the teacher signal and hinder knowledge transfer.

In this paper, we theoretically analyze the benefits of aggregation in FD and propose a particular method that improves model training under data heterogeneity.
Our contributions are summarized below. We...
\begin{enumerate}
\item \ldots introduce a family of uncertainty-aware approaches to aggregating predictions from different clients. See Section~\ref{sec:method}.
\item \ldots (under mild assumptions) we provide convergence guarantees for the two-stage training procedure: a nonconvex stationarity rate and (under a PL condition) a linear rate to a neighborhood whose size is determined by teacher quality. See Section~\ref{sec:theory}.
\item \ldots validate the proposed methods and baselines\citep{li2019fedmd,lin2020ensemble} on image and textual data under distributed heterogeneous scenarios. Empirically, in highly heterogeneous setups, our methods are the most robust. See Section~\ref{sec:experiments}.
\end{enumerate}

\section{Related Work}
\label{sec:related_work}

In this section, we discuss two major paradigms of distributed learning, Federated Learning and Federated Distillation, and then review uncertainty quantification methods that motivate our approach.

\paragraph{Federated Learning.}
Federated Learning (FL) was introduced by \citet{mcmahan2017fedavg} with the FedAvg algorithm.
In FL, distributed clients aim to train a global (shared) model collaboratively, but they are restricted from sharing their data. 
Instead, the clients are allowed to share gradients or model parameters with a server, which aggregates them across communication rounds. 
While FedAvg performs well when client data are homogeneous, its performance degrades significantly under data heterogeneity~\citep{li2020federated,karimireddy2020scaffold}.
Several methods address this through variance reduction, such as SCAFFOLD~\citep{karimireddy2020scaffold}, or through personalization~\citep{collins2021exploiting,kotelevskii2022fedpop,li2021ditto}.
However, FL requires all clients to share the same model architecture, which can be restrictive in practice and creates significant communication overhead for large models.

\paragraph{Federated Distillation.}
Federated Distillation (FD) was proposed to reduce communication costs and relax the shared-architecture constraint~\citep{li2019fedmd,seo202216,li2024federated,sattler2020communication}. 
FD draws on knowledge distillation~\citep{hinton2015distilling,gou2021knowledge}, in which a student model learns to match the teacher's predictions.
In the federated setting, the teacher signal is formed by aggregating predictions from multiple clients on a shared public dataset. 
The change from exchanging weights to predictions reduces communication overhead and naturally supports heterogeneous client architectures, and led to different methods~\citep{shao2024selective,wei2022knowledge,lin2020ensemble}. 

However, constructing a high-quality teacher signal from heterogeneous client data remains a key challenge for FD.
Under \emph{class mismatch}, an aggregation that averages predictions of different clients~\citep{li2019fedmd,lin2020ensemble} treats all client predictions equally, regardless of whether a given client has seen the relevant classes during training.
This can result in out-of-distribution predictions being weighted equally with informed ones, leading to an unreliable teacher signal. 
In our work, we directly address this issue by proposing aggregation strategies that account for client reliability.

\paragraph{Uncertainty Quantification.}
Uncertainty quantification provides tools for assessing the reliability of model predictions~\citep{hullermeier2021aleatoric,gawlikowski2023survey}.
A variety of approaches exist to quantify predictive uncertainty and improve the reliability of outputs. These includes ensemble-based methods~\citep{lakshminarayanan2017simple}, Bayesian approaches~\citep{blundell2015weight}, and post-hoc calibration techniques~\citep{guo2017calibration}.
In our setting, density-based methods are particularly relevant~\citep{van2020uncertainty,mukhoti2023deep,kotelevskii2022nonparametric}, as they can estimate epistemic uncertainty, which reflects a model's awareness of the data it predicts on, for a single deterministic model without requiring an ensemble.

We build on these density-based methods and incorporate each client's uncertainty estimate into the server-side aggregation step.
This allows the server to down-weight unreliable client predictions when constructing the teacher signal. 

In the next sections, we formalize our setting and analyze how aggregation quality controls the convergence neighborhood of the supervised objective.
We then derive a practical method that outperforms standard averaging under high data heterogeneity while matching it at low heterogeneity.

\section{Setting and Method}
\label{sec:method}

We start this section by formalizing our setup and introducing the necessary notation.
We consider \(M\) distributed clients, collaboratively solving a classification problem with \(C\) classes.
Each client \(i \in \mathcal{S} = \{1, \ldots, M\}\) holds a private labeled dataset \(\DC_i \sim \PDi\), where labels come from a subset of all possible classes \(\CC_i \subseteq \{1, \dots, C\}\).

Following standard restrictions of the federated setup, we assume that clients cannot share their \emph{private} data.
We also assume that all clients have access to a \emph{shared} public unlabeled dataset \(\Dpub \sim \PDpubx\), containing covariates from all classes, where \(\PDpubx\) is a marginal distribution of \(\mathcal{P}_{\text{pub}}\) w.r.t.\ the input \(x\) without label.
Below, all guarantees are stated with respect to \(x \sim \PDpubx\).

We assume that each client \(i\) has a local model parameterized by \(\theta^{(i)} \in \RR^d\) and aims to minimize its expected loss over \(\PDi\):
\begin{equation*}
F_i(\theta) \coloneqq \E_{\zeta \sim \PDi}[L(\theta, \zeta)],
\end{equation*}
where \(L(\theta, \zeta)\) is a proper point-wise loss (e.g., cross-entropy). The global objective is to minimize the average loss over clients:
\begin{equation*}
\min_{\theta \in \RR^d} F(\theta) \coloneqq \frac{1}{M}\sum_{i=1}^M F_i(\theta),
\quad
\theta^* \in \arg\min_{\theta \in \RR^d} F(\theta).
\end{equation*}
Let \(f(x, \theta) \in \Delta^{C-1}\) denote the predicted probability vector over \(C\) classes and let \(p^*(x) \coloneqq f(x, \theta^*)\) be the reference predictor.

In what follows, our theoretical analysis assumes all clients use models of the same size. In practice, however, clients can use different architectures, since logit-based methods only exchange predictions rather than model parameters. We do not cover this heterogeneous-architecture case in the theory.

\begin{figure}[t]
  \centering
  \includegraphics[width=\columnwidth]{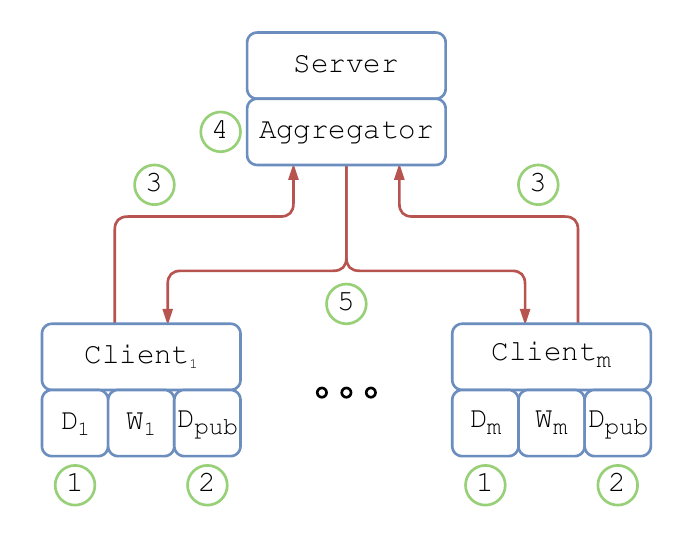}
  \caption{
  Our two-stage training loop works as follows.
  \textbf{Stage one.} (1) Clients train on their private labeled data.
  (2) Clients compute predictions on the public unlabeled dataset.
  (3) Clients send predictions to the server.\\
  \textbf{Stage two.} (4) The server aggregates predictions into soft labels.
  (5) The soft labels are sent back to clients for refinement.
      }
  \label{fig:overview}
\end{figure}

\subsection{Two-stage training}
\label{sec:setup}

A single training round consists of two stages (see Figure~\ref{fig:overview}).

In \textbf{Stage 1}, each client trains on its local data. It then computes predictions \(q_i(x) = \mathrm{softmax}(f_i(x; \theta_i)) \in \Delta^{C-1}\) for every \(x \in \Dpub\) and sends them to the server.

In \textbf{Stage 2}, the server forms a teacher \(\pagg(x) \in \Delta^{C-1}\) by aggregating the received predictions.
It then sends the aggregated vectors back to the clients, which refine their models on \(\Dpub\).

\begin{definition}\label{def:agg}
We say that the aggregation rule is probability mixing if there exist weights \(w_i(x) \ge 0\), \(\sum_{i \in \mathcal{S}} w_i(x) = 1\) such that
\begin{equation*}
\pagg(x) = \sum_{i \in \mathcal{S}} w_i(x) q_i(x).
\end{equation*}
\end{definition}
In practice, one may aggregate logits and then apply softmax. Our analysis is \emph{aggregation-agnostic} at the level of \(\pagg(x)\).
All rates depend on the teacher error \(\E_x \|\pagg(x) - \pstar(x)\|^2\).
Definition~\ref{def:agg} is used only when we want explicit \(M\)-dependence (e.g. \(1/M\)) from moment assumptions.
In the following sections, we discuss different aggregation strategies for the server to combine client predictions. We illustrate the setup in Figure~\ref{fig:overview} and summarize the procedure in Algorithm~\ref{alg:logit-fl}.
\begin{algorithm}[tb]
  \caption{Uncertainty-Aware Federated Distillation}
  \label{alg:logit-fl}
  \begin{algorithmic}[1]
    \REQUIRE Private datasets \(\{\DC_i\}_{i=1}^M\), public dataset \(\Dpub\),
    rounds \(R\), learning rates \(\eta, \eta_{pub}\), private epochs \(E_{priv}\), public epochs \(E_{pub}\),
    aggregation rule \(\mathsf{Agg}\),
    calibration datasets \(\{\DC_i^{cal}\}_{i=1}^M\)
    \ENSURE Client models \(\{f_i(\cdot;\theta_i)\}_{i=1}^M\)

    \STATE Initialize \(\{\theta_i^0\}_{i=1}^M\)
    \FOR{round \(r = 1\) to \(R\)}
      \FOR{each client \(i \in [M]\) \textbf{in parallel}}
        \STATE \textbf{Private training:} update \(\theta_i\) on \(\DC_i\) for \(E_{priv}\) epochs using cross-entropy
        \STATE \textbf{Public inference:} \\
         \(q_i(x) \gets \mathrm{softmax}(f_i(x;\theta_i))\) for all \(x \in \Dpub\)
        \STATE fit GMM \(G_i\) on \(\DC_i^{cal}\) 
        \STATE Upload \(G_i\) and \(\{q_i(x)\}_{x \in \Dpub}\) to the server
      \ENDFOR

      \FOR{each \(x \in \Dpub\)}
        \STATE Soft targets \(\pagg(x) \gets \mathsf{Agg}\big(\{(q_i(x), G_i)\}_{i=1}^M\big)\)
      \ENDFOR
      \STATE Broadcast \(\{\pagg(x)\}_{x \in \Dpub}\) to all clients

      \FOR{each client \(i \in [M]\) \textbf{in parallel}}
        \STATE \textbf{Public refinement:} update \(\theta_i\) on \(\Dpub\) for \(E_{pub}\) epochs using \(\CE\big(\pagg(x) \| \mathrm{softmax}(f_i(x;\theta_i))\big)\)
      \ENDFOR
    \ENDFOR
  \end{algorithmic}
\end{algorithm}

In the next section, we analyze the training procedure theoretically and show linear convergence to an optimal neighborhood that shrinks as more clients participate.


\section{Theoretical Analysis}
\label{sec:theory}


We provide a theoretical analysis of logit aggregation that justifies why averaging client predictions on a public dataset can recover an optimal predictor, and how aggregation of logits followed by finetuning on public data yields a provable improvement in the supervised objective.





\subsection{Teacher aggregation decomposition}

\begin{lemma}[Teacher MSE under bias-variance-correlation]
\label{lem:teacher_mse_weighted}
Consider probability-mixing aggregation $p_{\mathrm{agg}}(x)=\sum_{i\in\mathcal S} w_i(x)q_i(x)$ and assume the boundedness of its bias by $B_w$ as well as the boundedness of variance and correlation of $q_i(x) - p^*(x)$ by $\sigma^2$ and $\rho_c\sigma^2$ respectively (see Assumption~\ref{ass:teacher}).
Let
\[
\chi(x):=\sum_{i\in\mathcal S} w_i(x)^2 \geq \frac{1}{|\mathcal S|},\qquad \chi:=\E_x[\chi(x)].
\]
Then, the teacher MSE satisfies:
\begin{align}\label{eq:teacher_mse_bvc}
\varepsilon_{\mathrm{teach}}
&:=
\E_x\|p_{\mathrm{agg}}(x)-p^*(x)\|_2^2 \nonumber \\
&\le
B_w^2
\;+\;
\sigma^2\Big(\rho_c+(1-\rho_c)\chi\Big).
\end{align}
\end{lemma}

Lemma~\ref{lem:teacher_mse_weighted} decomposes the teacher MSE into a \emph{weighted bias} term $B_w^2$
and a \emph{variance/correlation} term controlled by the weight concentration $\chi=\E_x\sum_i w_i(x)^2$.
This makes the effect of the aggregation rule explicit.

\subsection{Averaging Aggregation}
\label{sec:avg}

The simplest existing approach is to average predictions across clients~\citep{li2019fedmd,lin2020ensemble}:
\begin{equation*}
\pagg^{\mathrm{AVG}}(x) = \frac{1}{M}\sum_{i=1}^M q_i(x).
\end{equation*}
Lemma~\ref{lem:teacher_mse_weighted} implies that $w_i(x)\equiv 1/M$ for all $i\in\mathcal S$ and all $x$, hence $\chi_{\mathrm{AVG}}=1/M$.
Thus, averaging yields the strongest possible variance reduction among all convex weights (it minimizes $\chi$ subject to $\sum_i w_i=1$).

However, the bias term becomes the bias of an \emph{unconditional mixture} of client predictors:
\[
B_{w,\mathrm{AVG}}^2=\E_x\big\|\E[\tfrac{1}{M}\sum_{i\in\mathcal S}\xi_i(x)\mid x, w]\big\|^2,
\]
so clients that are systematically unreliable on a given $x$ still contribute equally to the teacher.

While straightforward, this treats all clients equally informed and reliable, regardless of how informative they actually are. A client that is simply guessing will have the same weight as one producing meaningful predictions. Under the class-mismatch setup we consider, some clients have never seen covariates from certain classes in the public dataset, so their predictions on these objects can be misleading.

\subsection{Convergence rates}\label{sec:theory_rates_main}

The public distillation objective for any model parameters \(\theta\) is
\begin{equation*}
J_{\mathrm{agg}}(\theta) \coloneqq \E_{x \sim \PDpubx}\Big[\KL\big(\pagg(x) \| f(x, \theta)\big)\Big].
\end{equation*}

Stage~2 runs SGD on \(J_{\mathrm{agg}}\). 
At iteration \(t\), client $i$ draws \(\{x_{t,j}\}_{j=1}^b \stackrel{i.i.d.}{\sim} \PDpubx\) and sets:
\begin{align}\label{eq:sgd_step}
&g_t^{(i)} \coloneqq \frac{1}{b}\sum_{j=1}^b \nabla_\theta \KL\big(\pagg(x_{t,j}) \| f(x_{t,j}, \theta^{(i)}_t)\big), \nonumber\\
&\theta_{t+1}^{(i)} = \theta_t^{(i)} - \gamma g_t^{(i)},
\quad
\E[g_t^{(i)} \mid \theta_t^{(i)}] = \nabla J_{\mathrm{agg}}(\theta_t^{(i)}).
\end{align}
Since our analysis of Stage 2 applies to any client $i$, we omit the superscript $(i)$ in the formulas and results that follow.

\paragraph{Biased gradient oracle model.}
Although Stage~2 optimizes $\Jagg$, we evaluate progress in terms of the objective function $F$. Therefore, if we define
\[
\beta(\theta) := \nabla \Jagg(\theta)-\nabla F(\theta),
\quad
n_t := g_t-\nabla\Jagg(\theta_t),
\]
then we can interpret the procedure described in \eqref{eq:sgd_step} as biased SGD $\theta_{t+1} = \theta_t - \gamma g_t$ applied to the minimization of function $F$ with biased stochastic gradient:
\begin{equation*}
g_t = \nabla F(\theta_t) + \beta(\theta_t) + n_t,
\quad
\E[n_t\mid\theta_t]=0.
\end{equation*}

Under our aggregation and regularity assumptions (listed in Appendix~\ref{sec:assumptions}),
we show in Proposition~\ref{prop:implies-ajalloeian} that this bias satisfies an
\emph{expected biased-oracle} inequality of the form
$\E\|\beta(\theta_t)\|^2 \le m\,\E\|\nabla F(\theta_t)\|^2+\zeta^2$.
This is a weaker version of the pointwise oracle used by \citet{ajalloeian2021convergencesgdbiasedgradients}, and we therefore provide
an expected-oracle adaptation of their descent argument in Appendix~\ref{sec:convergence_rates_proof}.

\begin{proposition}[Expected-oracle conditions adaptation]
\label{prop:implies-ajalloeian}
Assume the boundedness of aggregation bias by $B_w$ as well as the boundedness of variance and correlation of $q_i(x) - p^*(x)$ by $\sigma^2$ and $\rho_c\sigma^2$ respectively (see Assumption~\ref{ass:teacher}); $c\in(0,1]$ and $b\ge 0$ quantify the alignment between $\nabla F(\theta_t)$ and the ideal distillation objective $\nabla \Jstar(\theta_t)$ (Assumption~\ref{ass:align}); $G_{\mathrm{sc}}^2$ bounds the expected squared Frobenius norm of the score matrix (Assumption~\ref{ass:score}); and $\nu^2$ bounds the variance of the stochastic gradient $g_t$ (Assumption~\ref{ass:sgd}). 
Fix $\alpha>\frac{c}{1-c}$ and define
\begin{align}\label{eq:mz_def_main}
m &:= \Big(1+\frac{1}{\alpha}\Big)c, \nonumber \\
\zeta^2 &:= (1+\alpha)G_{\mathrm{sc}}^2\,\varepsilon_{\mathrm{teach}}
+\Big(1+\frac{1}{\alpha}\Big)b^2\big.
\end{align}

Then the bias satisfies
\begin{equation}\label{eq:bias_bound_main}
\E\big[\|\beta(\theta_t)\|_2^2\big]
\le
m\,\E\big[\|\nabla F(\theta_t)\|_2^2\big] + \zeta^2,
\end{equation}
and the noise satisfies
\begin{equation}\label{eq:noise_bound_main}
\E\big[\|n_t\|_2^2\mid \theta_t\big]\le \nu^2.
\end{equation}
Therefore, the Stage~2 update \eqref{eq:sgd_step} fits the biased SGD oracle model of \citep{ajalloeian2021convergencesgdbiasedgradients}
(with $M_{\text{Aj}}=0$ and $\sigma^2_{\text{Aj}}=\nu^2$ in their Assumption~3).
\end{proposition}

Below we state the resulting nonconvex stationarity and PL convergence bounds, highlighting how the constants
$(m,\zeta^2,\nu^2)$ depend on the aggregation quality through $\varepsilon_{\mathrm{teach}}$
(and, via Lemma~\ref{lem:teacher_mse_weighted}, on the weight concentration $\chi$ and weighted bias $B_w^2$). We also translate these bounds into iteration complexities.

\begin{corollary}[Big-$\mathcal O$ complexity for nonconvex stationarity]
\label{cor:expected_oracle_bigO_nonconvex}
Assume F is $L_F$-smooth and let the conditions of Proposition~\ref{prop:implies-ajalloeian} hold. Define $m,\, \zeta,\,\nu^2$ as in that proposition. Let $\epsilon_{\min}:=\zeta^2/(1-m)$ and fix any $\epsilon>\epsilon_{\min}$.
Define $\bar\epsilon:=\epsilon-\epsilon_{\min}$.
Choosing $\gamma:=\min\{1/L_F,\ \bar\epsilon(1-m)/(2L_F\nu^2)\}$, it suffices to take
\[
T=\mathcal O\!\left(\frac{L_F\nu^2\,(F(\theta_0)-F_{\inf})}{(1-m)^2\,\bar\epsilon^2}\right)
\]
to ensure $\E\|\nabla F(\theta_{\mathrm{out}})\|^2\le \epsilon$.
\end{corollary}

\begin{corollary}[Big-$\mathcal O$ complexity under PL]
\label{cor:expected_oracle_bigO_pl}
Assume F is $L_F$-smooth and satisfies the Polyak–Łojasiewicz (PL) condition:$\|\nabla F(\theta)\|^2\ge 2\mu_F(F(\theta)-F^*),\, \mu_F>0$. Let the conditions of Proposition~\ref{prop:implies-ajalloeian} hold and define $m,\, \zeta,\,\nu^2$ as in that proposition.
Define the floor
\[
\epsilon_{\mathrm{floor}}(\gamma):=\frac{\zeta^2+L_F\gamma\nu^2}{2\mu_F(1-m)}.
\]
For any $\epsilon>\epsilon_{\mathrm{floor}}(\gamma), \, \gamma\le 1/L_F$, it suffices to take
\[
T=\mathcal O\!\left(\frac{1}{\gamma\mu_F(1-m)}\log\frac{F(\theta_0)-F^*}{\epsilon-\epsilon_{\mathrm{floor}}(\gamma)}\right)
\]
to ensure $\E[F(\theta_T)-F^*]\le \epsilon$.
\end{corollary}

\paragraph{Proof sketch (nonconvex stationarity).}
Stage~2 updates satisfy the biased-gradient decomposition $g_t=\nabla F(\theta_t)+\beta(\theta_t)+n_t$.
Proposition~\ref{prop:implies-ajalloeian} (Appendix) bounds the bias in expected-oracle form
$\E\|\beta(\theta_t)\|^2 \le m\,\E\|\nabla F(\theta_t)\|^2+\zeta^2$ and controls the noise by
$\E[\|n_t\|^2\mid\theta_t]\le \nu^2$.
Applying $L_F$-smoothness to one SGD step and taking conditional expectation yields a descent inequality with a negative
$\E\|\nabla F(\theta_t)\|^2$ term plus bias/noise corrections. Summing over $t$ and sampling
$\theta_{\mathrm{out}}$ uniformly from $\{\theta_0,\ldots,\theta_{T-1}\}$ gives the stated stationarity bound.
Full details are in Appendix~\ref{sec:theory:appendix}.

\paragraph{Proof sketch (PL linear rate).}
Starting from the same one-step descent inequality, the PL condition converts the gradient-norm term into a contraction in
$\E[F(\theta_t)-F^*]$, yielding a linear recursion with rate $1-\gamma\mu_F(1-m)$ and an additive floor proportional to
$(\zeta^2+L_F\gamma\nu^2)/(\mu_F(1-m))$. Unrolling the recursion gives the claimed linear convergence to a neighborhood.
Full details are in Appendix~\ref{sec:theory:appendix}.

\paragraph{On the error floor.}
Both results establish convergence only up to an irreducible optimization error. 
In the general nonconvex setting, this error is given by 
$\epsilon_{\min} := \zeta^2/(1-m)$, 
which is consistent with existing results for biased SGD \citep{ajalloeian2021convergencesgdbiasedgradients}. 
However, as shown in Proposition~\ref{prop:implies-ajalloeian}, the error floor can be reduced through improved aggregation. Specifically, the first term in the definition of $\zeta^2$ in \eqref{eq:mz_def_main} is proportional to 
$\varepsilon_{\mathrm{teach}} := \E_x\|p_{\mathrm{agg}}(x)-p^*(x)\|_2^2$, 
which quantifies the quality  of the aggregation rule. 
In the case of (near-)ideal aggregation, we have 
$\varepsilon_{\mathrm{teach}} \approx 0$, 
and consequently $\zeta \sim b^2$, 
where $b$ (together with $c$) characterizes the alignment between $\nabla F(\theta_t)$ and the ideal distillation objective $\nabla \Jstar(\theta_t)$ (Assumption~\ref{ass:align}). 
As discussed in the appendix (see Remark~\ref{rem:on_grad_align}), it is natural to expect $b$ to be small toward the end of training. 
Therefore, our results imply convergence to a small and practically meaningful optimization error, provided that the aggregation rule is sufficiently accurate.


\paragraph{Regarding single-round scope.}
Theorems in Section~\ref{sec:theory_rates_main} analyze a single Stage~2 distillation phase with a fixed teacher
$p_{\mathrm{agg}}$ (computed once from Stage~1 predictors). This isolates the effect of aggregation quality through
$\varepsilon_{\mathrm{teach}}$ (and $\chi$) on the optimization floor.

In a multi-round procedure, each round recomputes a teacher $p_{\mathrm{agg}}^{(r)}$ and runs Stage~2 for $T_2$ steps.
The single-round analysis applies within each round with round-dependent quantities
$\varepsilon_{\mathrm{teach},r}$, $\chi_r$, and the induced oracle parameters $(m_r,\zeta_r^2)$.
Without additional restrictions on Stage~1, monotone improvement across rounds cannot be guaranteed in the worst case. However, if teacher-quality proxies (e.g., $\varepsilon_{\mathrm{teach},r}$ or $\chi_r$) improve over rounds,
the theory predicts a shrinking neighborhood floor accordingly.

\paragraph{Proofs and further details.}
All missing technical details and complete proofs are provided in Appendix~\ref{sec:theory:appendix}.

\subsection{UWA: Uncertainty-Weighted Averaging}
\label{sec:uwa}

To address issues raised in Section~\ref{sec:avg}, we propose weighting each client's prediction by its confidence.
We use a density-based approach that links predictive uncertainty to how typical the model's outputs look~\citep{van2020uncertainty,mukhoti2023deep,kotelevskii2022nonparametric}.

For each client \( i \), we keep a separate labeled calibration split \( \DC_i^{cal} \) (20\% of private data in our experiments) that is not used for training, but is used to fit a density model.
There are multiple options for selecting a density model, depending on the level of flexibility required. One of the reasonable choices is a Gaussian Mixture Model (GMM) with \( |\CC_i| \) components (one per class \( k \in \CC_i \)), equal mixture weights, and diagonal covariance.
We fit the GMM to the logits produced on the calibration set \( \DC_i^{cal} \): \( z_i(x) = f_i(x; \theta_i) \in \RR^C \). We refit it after each training phase on a local (private) dataset.

Let us now describe how we use it for objects coming from the shared public dataset.
Denote by \( \mu_{i,k,d} \) and \( \sigma_{i,k,d} \) the mean and standard deviation of the \( d \)-th logit component for class \( k \) on client \( i \).
To evaluate the reliability of a client, we compute the log-likelihood under the fitted GMM for each \( x \) from the shared public dataset:
\begin{align*}
\ell_i(x) &\coloneqq \log p(x \mid i) \\
&= \log\Bigl(\frac{1}{|\CC_i|}\sum_{k \in \CC_i} \mathcal{N}\bigl(z_i(x); \mu_{i,k}, \sigma^2_{i,k}\bigr)\Bigr),
\label{eq:gmm_score}
\end{align*}
where \( \mathcal{N}(\cdot; \mu, \Sigma) \) denotes a Gaussian with mean \( \mu \) and covariance \( \Sigma \).

The value of \( \ell_i(x) \) tells us how reliable the prediction of client \( i \) is for input \( x \).
When \( \ell_i(x) \) is large, the logits that the client predicts look similar to what it saw during training, so it is likely producing informed predictions.
When \( \ell_i(x) \) is small, the logits are unusual for this client, which typically happens for classes outside \( \CC_i \).
Therefore, the predictions are likely not well-informed, and the client is instead guessing.

We then produce weights from these log-likelihood values by applying a softmax over all clients for a fixed input \( x \):
\begin{equation*}
w_i(x) = \frac{\exp(\ell_i(x))}{\sum_{j=1}^M \exp(\ell_j(x))} = \frac{p(x \mid i)}{\sum_j p(x \mid j)}.
\label{eq:uwa_weights}
\end{equation*}
Note that if \( p(i) = p(j) \) for all \( i, j \), Bayes' theorem gives \( w_i(x) = p(i \mid x) \).

The soft target labels are then a weighted average of client predictions:
\begin{equation}
\pagg^{UWA}(x) = \sum_{i=1}^M w_i(x) q_i(x) \approx \E_{w_i(x)} q_i(x).
\label{eq:uwa}
\end{equation}

For UWA, the weights $w_i(x)$ depend on the input $x$, so $\chi(x)$ is generally \emph{larger} than $1/M$ when the weights concentrate on a subset of clients (less averaging):
\[
\chi_{\mathrm{UWA}}=\E_x\sum_{i\in\mathcal S} w_i(x)^2 \in \Big[\frac{1}{M},1\Big].
\]
This increases the variance term in Lemma~\ref{lem:teacher_mse_weighted} relative to AVG whenever weights are concentrated.
However, a data-dependent rule, such as UWA, may reduce $B_w^2$ by down-weighting clients whose conditional mean error is large, i.e. clients that are out-of-distribution for the given input. While logit-density scores have been used for out-of-distribution detection~\citep{mukhoti2023deep}, we adapt them here to federated aggregation under client heterogeneity.

More explicitly, let $\mu_i(x,w):=\E[\xi_i(x)\mid x,w(x)]$. Then
\begin{align*}
\E[\bar\xi(x)\mid x,w] &= \sum_{i\in\mathcal S} w_i(x)\mu_i(x,w), \\
\big\|\E[\bar\xi(x)\mid x,w]\big\|_2^2
&\le \sum_{i\in\mathcal S} w_i(x)\,\|\mu_i(x,w)\|_2^2,
\end{align*}
so the bias contribution can decrease when weights are concentrated on clients with smaller $\|\mu_i(x,w)\|_2$.

Hence, UWA trades off less variance reduction (larger $\chi$) for potentially smaller bias (smaller $B_w^2$). In regimes with strong client heterogeneity (large systematic bias), UWA can reduce $B_w^2$ substantially and outperform averaging. In more homogeneous regimes (small bias), uniform averaging tends to be preferable because it minimizes $\chi$.

\begin{table*}[t]
  \centering
  \caption{Best test accuracy (\%) across datasets and heterogeneity levels (\(k\) classes per client). Mean \(\pm\) std over 3 seeds. Best result per column in \textbf{bold}. \(Ref.\) is fully informed reference trained on all-classes dataset with size \(|\DC_{pub}| + |\DC_{i}|\)}
  \label{tab:main-results}
  \begin{tabular}{cl cccccc c}
    \toprule
    Dataset & Method & $k{=}2$ & $k{=}3$ & $k{=}4$ & $k{=}5$ & $k{=}7$ & $k{=}9$ & $Ref.$ \\
    \midrule
    \multirow{3}{*}{\textbf{CIFAR-10}}
      & AVG  & $21.70{\scriptstyle\pm 2.15}$ & $41.88{\scriptstyle\pm 1.73}$ & $56.79{\scriptstyle\pm 0.61}$ & $65.33{\scriptstyle\pm 2.65}$ & $72.55{\scriptstyle\pm 0.96}$ & $\mathbf{75.42}{\scriptstyle\pm 0.05}$ & \multirow{3}{*}{$80.84$} \\
      & UWA  & $\mathbf{39.51}{\scriptstyle\pm 2.69}$ & $59.37{\scriptstyle\pm 1.57}$ & $64.71{\scriptstyle\pm 0.49}$ & $68.96{\scriptstyle\pm 0.40}$ & $71.23{\scriptstyle\pm 0.15}$ & $73.16{\scriptstyle\pm 0.13}$ & \\
      & sUWA & $38.33{\scriptstyle\pm 0.63}$ & $\mathbf{60.96}{\scriptstyle\pm 1.42}$ & $\mathbf{67.45}{\scriptstyle\pm 1.17}$ & $\mathbf{71.38}{\scriptstyle\pm 0.27}$ & $\mathbf{73.94}{\scriptstyle\pm 0.21}$ & $74.98{\scriptstyle\pm 0.21}$ & \\
    \midrule
    \multirow{3}{*}{\textbf{Yahoo}}
      & AVG  & $49.61{\scriptstyle\pm 3.63}$ & $55.78{\scriptstyle\pm 1.76}$ & $56.39{\scriptstyle\pm 2.45}$ & $58.94{\scriptstyle\pm 1.48}$ & $61.17{\scriptstyle\pm 0.23}$ & $61.92{\scriptstyle\pm 0.40}$ & \multirow{3}{*}{$63.80$} \\
      & UWA  & $56.46{\scriptstyle\pm 1.06}$ & $57.90{\scriptstyle\pm 0.77}$ & $58.51{\scriptstyle\pm 0.16}$ & $59.69{\scriptstyle\pm 0.27}$ & $61.20{\scriptstyle\pm 0.45}$ & $61.97{\scriptstyle\pm 0.47}$ & \\
      & sUWA & $\mathbf{56.66}{\scriptstyle\pm 1.71}$ & $\mathbf{59.61}{\scriptstyle\pm 1.91}$ & $\mathbf{58.81}{\scriptstyle\pm 0.80}$ & $\mathbf{60.03}{\scriptstyle\pm 0.81}$ & $\mathbf{61.55}{\scriptstyle\pm 0.27}$ & $\mathbf{62.08}{\scriptstyle\pm 0.41}$ & \\
    \midrule
    & & $k{=}20$ & $k{=}30$ & $k{=}40$ & $k{=}50$ & $k{=}70$ & $k{=}90$ & \\
    \cmidrule(lr){3-8}
    \multirow{3}{*}{\textbf{CIFAR-100}}
      & AVG  & $52.73{\scriptstyle\pm 1.08}$ & $59.89{\scriptstyle\pm 0.34}$ & $64.02{\scriptstyle\pm 0.59}$ & $\mathbf{66.79}{\scriptstyle\pm 0.59}$ & $\mathbf{69.07}{\scriptstyle\pm 0.40}$ & $\mathbf{69.43}{\scriptstyle\pm 0.18}$ & \multirow{3}{*}{$69.95$} \\
      & UWA  & $59.82{\scriptstyle\pm 0.33}$ & $62.09{\scriptstyle\pm 0.60}$ & $63.43{\scriptstyle\pm 0.62}$ & $64.91{\scriptstyle\pm 0.79}$ & $66.76{\scriptstyle\pm 0.53}$ & $67.67{\scriptstyle\pm 0.09}$ & \\
      & sUWA & $\mathbf{60.85}{\scriptstyle\pm 0.31}$ & $\mathbf{63.29}{\scriptstyle\pm 0.39}$ & $\mathbf{64.81}{\scriptstyle\pm 0.50}$ & $66.20{\scriptstyle\pm 0.60}$ & $67.90{\scriptstyle\pm 0.39}$ & $68.35{\scriptstyle\pm 0.03}$ & \\
    \bottomrule
  \end{tabular}
\end{table*}

\paragraph{Smoothed UWA}
In our experiments, we observed that UWA can produce highly peaked weight distributions, especially in early rounds of training. 
We hypothesize that this is due to unreliable logits learned by clients, which result in flawed density models fitted on them.
This effect can arise from overfitting of clients' models to their local data.

In what follows, we present a practical modification of UWA that empirically improves performance.
We refer to it as \textbf{smoothed UWA (sUWA)}.

The idea is to scale the log-likelihoods produced by local density models in a temperature scaling-like manner~\citep{guo2017calibration}.
Specifically, we introduce a scalar temperature \(\tau > 0\) and produce sUWA weights as follows:
\begin{equation*}
w_i^{\mathrm{sUWA}}(x) = \frac{\exp(\tau \ell_i(x))}{\sum_{j \in \mathcal{S}} \exp(\tau \ell_j(x))}.
\end{equation*}

The method interpolates between three extremes:
\begin{itemize}
\item \( \tau = 1 \) recovers UWA; 
\item \( \tau \to \infty \) selects the most confident model;
\item \( \tau = 0 \) recovers AVG.
\end{itemize}

In our experiments, we set \( \tau < 1 \) to reduce the concentration of aggregation weight.
Specifically, we fix \( \tau = 0.25 \) across all experiments.

From the perspective of Lemma~\ref{lem:teacher_mse_weighted}, the temperature controls the bias-variance tradeoff: lower \(\tau\) decreases weight concentration \(\chi\) (better variance reduction) at the cost of higher bias \(B_w^2\).

\section{Experiments}
\label{sec:experiments}

In this section, we empirically study different aggregation approaches.
We consider classification in two domains:
(i) image classification on CIFAR-10~\citep{krizhevsky2009cifar10} trained from scratch,
(ii) image classification on CIFAR-100 using transfer learning~\citep{krizhevsky2009cifar10}, and
(iii) text classification on Yahoo Answers~\citep{chang2008importance} with transfer learning.
In all settings, we use \(M = 20\) clients and vary the degree of heterogeneity by changing the number of classes \(k\) assigned to each client.

We compare three aggregation methods, namely \textbf{AVG} (uniform averaging, baseline), \textbf{UWA} (uncertainty-weighted averaging), and \textbf{sUWA} (smoothed UWA with temperature \(\tau = 0.25\)).

\subsection{Experimental setup}
\label{sec:exp_setup}

Depending on the dataset, we consider different schemes of class distributions over clients:
\begin{itemize}
\item For \textbf{CIFAR-10}, each client has \(k \in \{2, 3, 4, 5, 7, 9\}\) classes. For the classification, we use ResNet-18~\citep{he2016deep} trained from scratch.
\item For \textbf{CIFAR-100}, \(k \in \{20, 30, 40, 50, 70, 90\}\) classes per client, and as a model we use ResNet-18, pretrained on ImageNet~\citep{deng2009imagenet}.
\item For \textbf{Yahoo Answers}~\citep{zhang2015character}, a topic classification benchmark with 10 classes, each client is assigned \(k \in \{2, 3, 4, 5, 7, 9\}\) classes. We use BERT-tiny~\citep{turc2019well}, pretrained on Wikipedia and BookCorpus, as the client model.
\end{itemize}

\subsection{Homogeneous setting}
\label{sec:sanity_check}

Before studying heterogeneous settings, we verify our theoretical analysis in the homogeneous case.
We train \(M = 20\) clients on CIFAR-10, each observing all 10 classes with 100 samples per class (1,000 private training samples per client).
The shared public dataset contains 500 samples per class (5,000 total).

We observe that communication between clients helps significantly.
Without it, local training achieves only \(42.40\% \pm 0.4\%\) test accuracy.
A single round of distillation already raises accuracy to \(58.04\% \pm 0.06\%\).
This confirms that aggregating clients' predictions improves convergence (see Section~\ref{sec:theory}), and unlabeled public data provides an effective channel for knowledge transfer. 

\begin{figure*}[t]
  \centering
  \includegraphics[width=\textwidth]{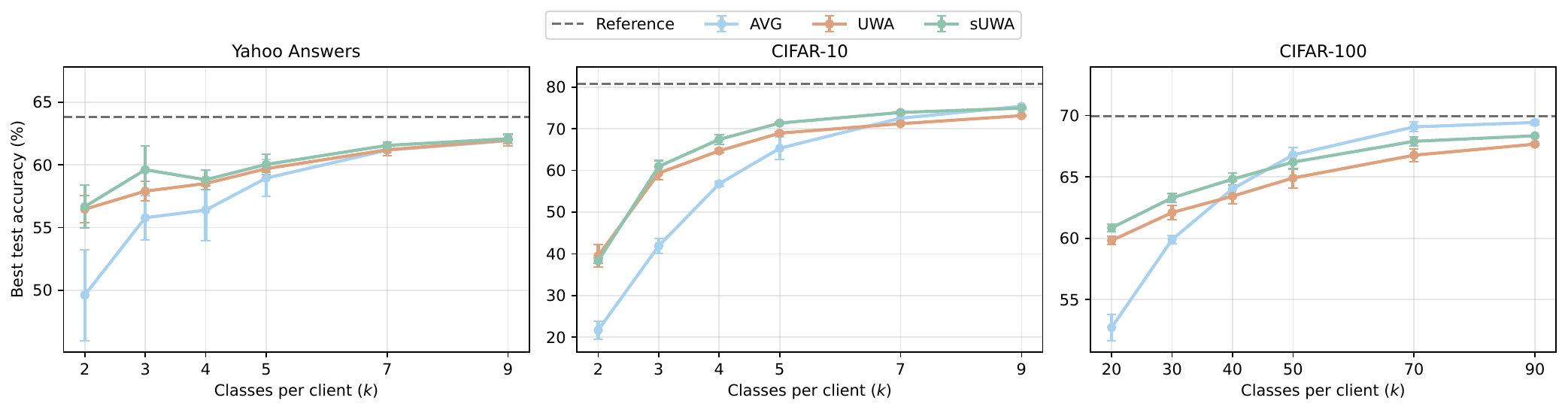}
  \caption{Best test accuracy vs.\ classes per client (\(k\)). Dashed line: fully-informed reference model trained on all-classes dataset with size \(|\DC_{pub}| + |\DC_{i}|\)).}
  \label{fig:acc_vs_k}
\end{figure*}

\subsection{Heterogeneous setting}
\label{sec:results}

In this section, we consider a practical scenario of distributed heterogeneous clients.

We present average client test accuracy across all methods and heterogeneity levels. The results are summarized in Table~\ref{tab:main-results}.
The density-weighted methods provide the most significant improvement at low \(k\) (i.e., high heterogeneity), where most public samples are out-of-distribution for a given client.
As heterogeneity decreases, clients become equally aware of all classes, and therefore UWA and sUWA converge to AVG. Without heterogeneity, all methods achieve nearly the same test accuracy.
Smoothed UWA consistently outperforms UWA, especially at moderate heterogeneity.

In Figure~\ref{fig:acc_vs_k}, we display the same results visually. We see that uncertainty-aware approaches (both UWA and sUWA) are most effective in high-data-heterogeneity settings.
On CIFAR-10, these approaches reach almost 40\% accuracy when clients observe only 20\% of classes.
As \(k\) increases, client predictions become more reliable, the weights in uncertainty-aware methods approach uniformity, reducing these methods to simple averaging.

Smoothed UWA tends to outperform both AVG and UWA across a broad range of \(k\).
We have the following explanation for this observation.
Early in training, local models can overfit to their private data, leading to unreliable density estimates.
This issue makes the uncertainty weights flawed and overly concentrated on a few clients.
The temperature parameter flattens the weight distribution, preventing any single client from dominating the aggregation (in practice, \( \tau = 0.25 \)).

\subsection{Communication cost}
\label{sec:comm_cost}

To demonstrate additional benefits of Federated Distillation, we compute the communication cost between clients and the server. 
We compare the cost of our logit-based methods against \texttt{SCAFFOLD}~\citep{karimireddy2020scaffold}, a gradient-based federated learning method that uses control variates to correct client drift.

Although \texttt{SCAFFOLD} converges in as few as 5 rounds while our methods require 20 to 30, the total bytes transferred are far lower for logit-based approaches.
On Yahoo Answers (BERT-tiny, \({\sim}4.3\)M parameters), \texttt{SCAFFOLD} transmits about 172MB per client over 5 rounds. 
In contrast, AVG and UWA send only about 2MB over 25 rounds (2,000 public samples \(\times\) 10 classes \(\times\) 4 bytes \(\times\) 25 rounds). 
This is an 86\(\times\) reduction.
On CIFAR-100 (ResNet-18, \({\sim}11.7\)M parameters, \(C = 100\), 5,000 public samples), the logit cost rises to 50MB over 25 rounds, while \texttt{SCAFFOLD} requires 468MB over 5 rounds, still a 9\(\times\) gap.

In terms of accuracy, at low heterogeneity (\(k \leq 4\)), \texttt{SCAFFOLD} performs comparably or marginally better.
At higher heterogeneity (\(k \geq 5\)), our methods match or surpass it while transferring orders of magnitude less data.

\begin{figure}[t]
  \centering
  \includegraphics[width=\columnwidth]{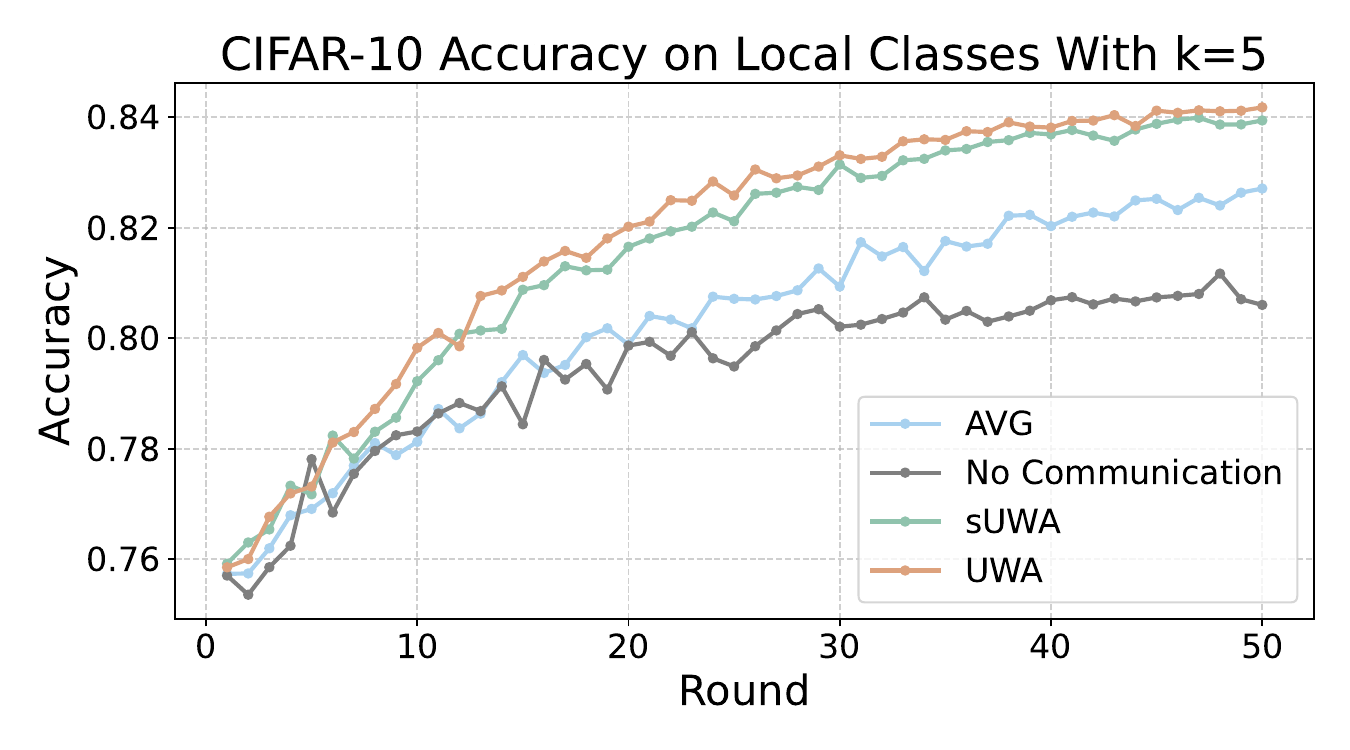}
  \caption{Mean accuracy on local classes over communication rounds.}
  \label{fig:local_accuracy}
\end{figure}

\subsection{Local class improvement}
\label{sec:local_class}

Over communication rounds, the test accuracy on local classes (measured after the local stage) improves significantly compared to training without communication (see Figure~\ref{fig:local_accuracy}).
The overlap between local class sets across clients can explain this improvement.
Our methods perform significantly better in this scenario as well.
%

\subsection{Limitations}
\label{sec:limitations}

Our approach relies on a shared, unlabeled public dataset accessible to all clients and the server.
While such data is available in many practical settings (e.g., publicly crawled text or images), it may not always be available in specialized domains, such as medical or financial data.
Additionally, the communication advantage over gradient-based methods decreases as the number of classes grows.
Each client sends a \((C-1)\)-dimensional probability vector per public sample, so for datasets with many classes and large public sets, the per-round cost increases accordingly, though it remains below gradient-based costs.

\section{Conclusion}
\label{sec:conclusion}
In this paper, we studied the problem of aggregating client predictions in Federated Distillation under data heterogeneity.

We provided a theoretical analysis showing that aggregating predictions from different clients on a shared public dataset converges to a neighborhood of the optimum, with the neighborhood size controlled by the quality of the aggregating strategy.

As specific aggregation strategies, we proposed UWA and sUWA, uncertainty-aware methods that downweight unreliable client predictions using the density of predicted logits.

In our experiments on image (CIFAR-10, CIFAR-100) and text (Yahoo Answers) datasets, we demonstrated that these methods are most effective in high-data-heterogeneity scenarios. At the same time, they match the standard averaging procedure in low- or no-heterogeneity settings.

\bibliography{sample}

\newpage
\appendix
\onecolumn

\title{Supplementary Material}
\maketitle
\section{Experimental Details}

\subsection{Hyperparameters}
\label{sec:hyperparameters}

\Cref{tab:hyperparams} summarizes the hyperparameters used across all experiments.

\begin{table}[H]
  \centering
  \caption{Hyperparameters for all experiments. All experiments use $M{=}20$ clients, 3 random seeds, and sUWA temperature $\tau{=}0.25$. For UWA and sUWA GMMs are fitted locally using the same private dataset, but augmented.} 
  \label{tab:hyperparams}
  \small
  \begin{tabular}{l ccc}
    \toprule
    \textbf{Parameter} & \textbf{Yahoo Answers} & \textbf{CIFAR-10} & \textbf{CIFAR-100} \\
    \midrule
    \multicolumn{4}{l}{\textit{Federation Setup}} \\
    Classes per client ($k$) & \{2,3,4,5,7,9\} & \{2,3,4,5,7,9\} & \{20,30,40,50,70,90\} \\
    Private samples/client & 1,500 & 5,000 & 5,000 \\
    Public dataset size & 2,000 & 5,000 & 5,000 \\
    Communication rounds & 50 & 50 & 50 \\
    \midrule
    \multicolumn{4}{l}{\textit{Model Architecture}} \\
    Backbone & BERT-tiny & ResNet-18 & ResNet-18 \\
    Pretrained & Yes (Wikipedia + BookCorpus) & No & Yes (ImageNet-1K) \\
    \midrule
    \multicolumn{4}{l}{\textit{Training}} \\
    Optimizer & AdamW & Adam & Adam \\
    Learning rate & $2{\times}10^{-5}$ & $1{\times}10^{-3}$ & $1{\times}10^{-4}$ \\
    Batch size & 32 & 128 & 128 \\
    First round epochs & 3 & 20 & 1 \\
    Subsequent epochs & 1 & 2 & 1 \\
    \bottomrule
  \end{tabular}
\end{table}

\subsection{Dataset Details}

\paragraph{Yahoo Answers.}
Yahoo Answers~\citep{zhang2015character} is a topic classification dataset with 10 classes (e.g., Science \& Mathematics, Sports, Health). It contains 1,400,000 training samples and 60,000 test samples. Each sample consists of a question title, question content, and best answer concatenated together.

\paragraph{CIFAR-10 and CIFAR-100.}
CIFAR-10 contains 60,000 $32 \times 32$ color images across 10 classes (50,000 train, 10,000 test). CIFAR-100 contains the same number of images but with 100 fine-grained classes grouped into 20 superclasses.

\subsection{Aggregation Method Details}

\paragraph{Simple Averaging (\texttt{AVG}).}
Computes the element-wise mean of client softmax probabilities:
\[
\bar{q}(x) = \frac{1}{M}\sum_{i=1}^M q_i(x), \quad q_i(x) = \text{softmax}(f_i(x)).
\]

\paragraph{UWA (Uncertainty-Weighted Averaging).}
Each client fits a Gaussian mixture model over its logit distribution using private training data (one component per local class). At aggregation time, the confidence score for client $i$ on input $x$ is:
\[
\ell_i(x) = \log p_{\text{GMM}_i}(f_i(x)), \quad w_i(x) = \frac{\exp(\ell_i(x))}{\sum_j \exp(\ell_j(x))}.
\]

\paragraph{sUWA (Smoothed UWA).}
Uses a temperature parameter $\tau = 0.25$ to smooth the UWA weights:
\[
w_i^{\text{sUWA}}(x) = \frac{\exp(\tau \cdot \ell_i(x))}{\sum_j \exp(\tau \cdot \ell_j(x))}.
\]
As $\tau \to 0$, this reduces to uniform averaging (\texttt{AVG}).

\subsection{UWA Confidence Score Formula}
\label{sec:uwa-formula}

For the Uncertainty-Weighted Averaging (UWA) method, the confidence score for client \(i\) on input \(x\) is computed as:
\begin{equation}
  \ell_i(x) \coloneqq \log\left(
    \frac{1}{|\CC_i|}
    \sum_{k \in \CC_i}
    (2\pi)^{-C/2}
    \prod_{d=1}^{C}\sigma_{i,k,d}^{-1}
    \exp\left[
      -\tfrac{1}{2}\sum_{d=1}^{C}
      \frac{\big(f_{i,d}(x)-\mu_{i,k,d}\big)^2}{\sigma_{i,k,d}^{2}}
    \right]
  \right),
\end{equation}
where we assume a mean-field approximation to each component and uniform component weights. The parameters \(\mu_{i,k,d}\) and \(\sigma_{i,k,d}\) are the mean and standard deviation of the \(d\)-th logit dimension for the \(k\)-th class component in client \(i\)'s Gaussian mixture model.

\section{Additional Results}

\subsection{Accuracy Curves}

\begin{figure}[H]
  \centering
  \includegraphics[width=\textwidth]{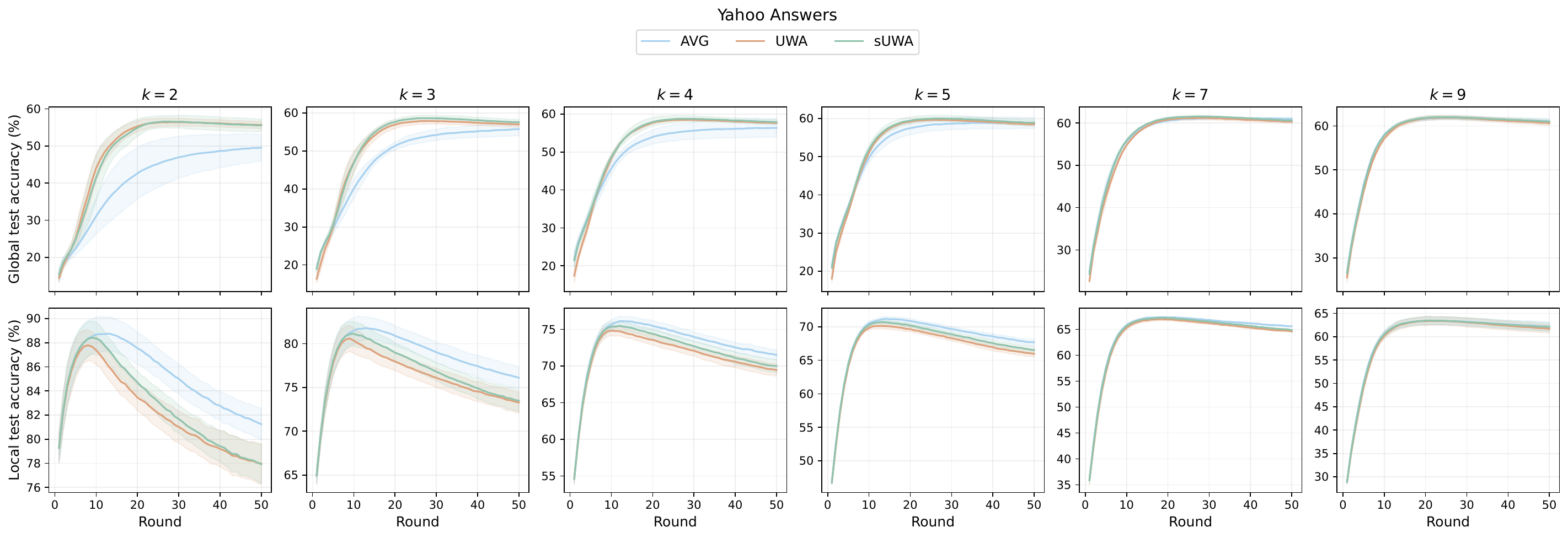}
  \caption{Yahoo Answers: global (top) and local (bottom) test accuracy across communication rounds for $k \in \{2,3,4,5,7,9\}$.}
  \label{fig:curves-yahoo}
\end{figure}

\begin{figure}[H]
  \centering
  \includegraphics[width=\textwidth]{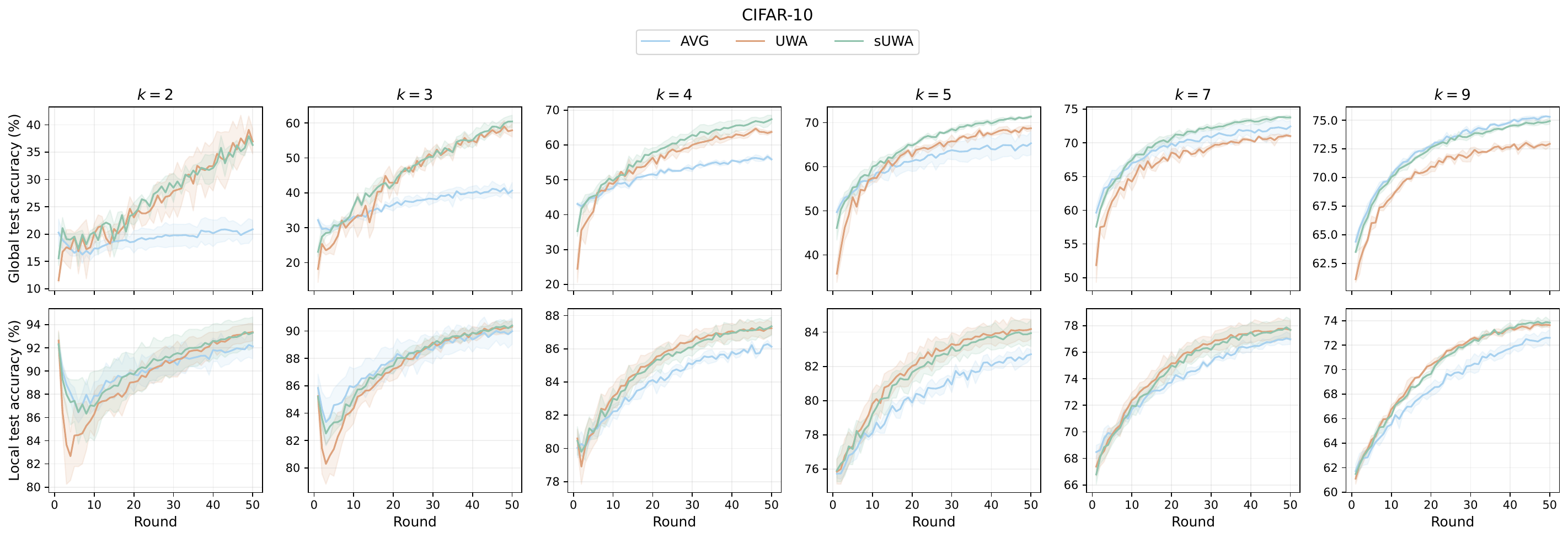}
  \caption{CIFAR-10: global (top) and local (bottom) test accuracy across communication rounds for $k \in \{2,3,4,5,7,9\}$.}
  \label{fig:curves-cifar10}
\end{figure}

\begin{figure}[H]
  \centering
  \includegraphics[width=\textwidth]{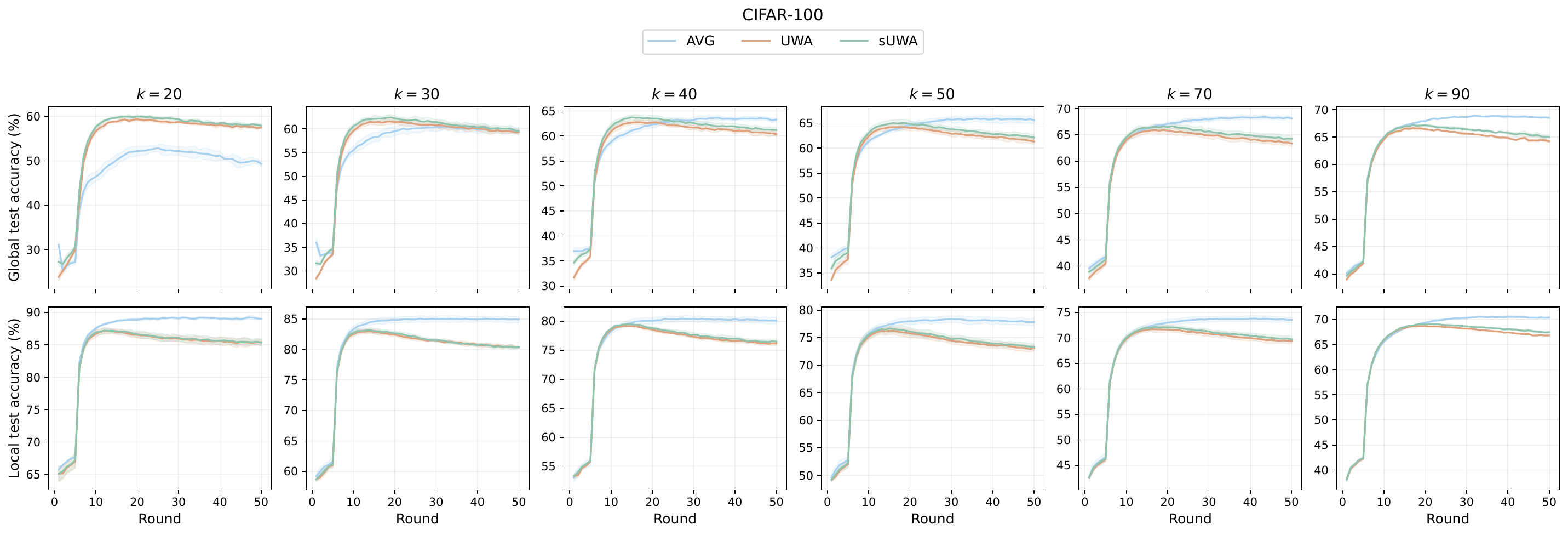}
  \caption{CIFAR-100: global (top) and local (bottom) test accuracy across communication rounds for $k \in \{20,30,40,50,70,90\}$.}
  \label{fig:curves-cifar100}
\end{figure}

\Cref{fig:curves-yahoo,fig:curves-cifar10,fig:curves-cifar100} show the global and local test accuracy across communication rounds for all datasets and heterogeneity levels. In each figure, the top row shows global test accuracy (evaluated on all classes) and the bottom row shows local test accuracy (evaluated only on each client's assigned classes). Lines show the mean over 3 seeds; shaded regions indicate $\pm 1$ standard deviation.

\subsection{Ablation: Public vs.\ Private Dataset Size}
\label{sec:ablation}

\begin{figure}[H]
  \begin{minipage}[t]{0.6\textwidth}
    \vspace{0pt}
    To understand how the sizes of the public and private datasets affect performance, we run an ablation study on Yahoo Answers using \texttt{sUWA} with $k{=}3$ and 30 communication rounds. We vary the public dataset size over $\{500, 1{,}000, 2{,}000, 5{,}000\}$ and private samples per client over $\{500, 1{,}000, 1{,}500, 3{,}000\}$.

    \Cref{fig:ablation-yahoo} shows the results as a heatmap. Increasing the public dataset size has the largest impact on accuracy: moving from 500 to 2,000 public samples yields a ${\sim}17$ percentage-point improvement regardless of private data size. Beyond 2,000 public samples, gains saturate unless private data also increases. Private data size has a smaller but consistent effect, especially when public data is abundant. This suggests that under a fixed data budget, prioritizing the collection of shared public data yields larger returns than expanding per-client private data, since the public set mediates all inter-client knowledge transfer in the distillation framework.
  \end{minipage}%
  \hfill
  \begin{minipage}[t]{0.35\textwidth}
    \vspace{0pt}
    \centering
    \includegraphics[width=\textwidth]{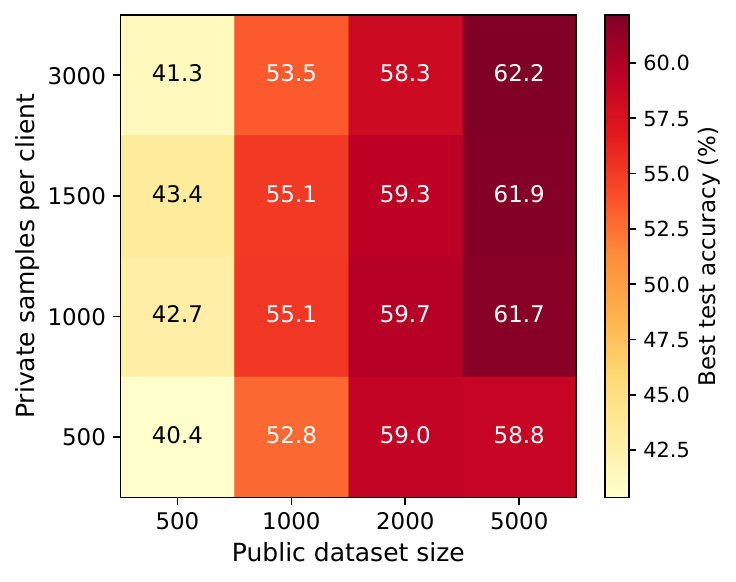}
    \captionof{figure}{Ablation on Yahoo Answers (\texttt{sUWA}, $k{=}3$): best test accuracy (\%) as a function of public dataset size and private samples per client.}
    \label{fig:ablation-yahoo}
  \end{minipage}
\end{figure}

\section{Theoretical Analysis (Lemmas and Proofs)}
\label{sec:theory:appendix}
In this section, we provide the missing details on the theoretical convergence analysis sketched in Section~\ref{sec:theory}.

\subsection{Notation Table}

\begin{table}[h]
\centering
\small
\begin{tabular}{ll}
\toprule
Symbol & Meaning \\
\midrule
$\mathcal{S},M$ & participating client set and its size ($|\mathcal{S}|=M$) \\
$C$ & number of classes \\
$\theta\in\RR^d$ & model parameters \\
$f(x,\theta)\in\Delta^{C-1}$ & model predictive probability vector \\
$F(\theta)=\frac1M\sum_i F_i(\theta)$ & global supervised objective \\
$F^*=\inf_\theta F(\theta)$ & optimal value of $F$ \\
$\theta^*$ & a minimizer of $F$ \\
$\pstar(x)=f(x,\theta^*)$ & reference predictor at optimum \\
$\Dpubx$ & public input distribution \\
$q_i(x)=f(x,\theta_i)$ & client predictor after Stage~1 \\
$\pagg(x)$ & aggregated teacher distribution \\
$w_i(x)$ & aggregation weights ($w_i\ge 0$, $\sum_i w_i=1$) \\
$\Jagg(\theta)$ & Stage~2 distillation objective w.r.t.\ $\pagg$ \\
$\Jstar(\theta)$ & ideal KD objective w.r.t.\ $\pstar$ \\
$g_t$ & stochastic gradient estimator for $\nabla\Jagg(\theta_t)$ \\
$\nu^2$ & bound on Stage~2 gradient noise variance (Assumption~\ref{ass:sgd}) \\
$(B,\sigma,\rho_c)$ & aggregation bias/variance/correlation constants (Assumption~\ref{ass:teacher}) \\
$c,b$ & alignment constants (Assumption~\ref{ass:align}) \\
$G_{\mathrm{sc}}^2$ & score-energy bound (Assumption~\ref{ass:score}) \\
$\varepsilon_{\mathrm{teach}}$ & teacher MSE $\E_x\|\pagg(x)-\pstar(x)\|^2$ \\
$\beta(\theta)$ & Stage~2 bias $\nabla\Jagg(\theta)-\nabla F(\theta)$ \\
$n_t$ & zero-mean noise $g_t-\nabla\Jagg(\theta_t)$ \\
$m,\zeta^2$ & biased-oracle constants (Prop.~\ref{prop:implies-ajalloeian}) \\
\bottomrule
\end{tabular}
\end{table}

\subsection{Assumptions}\label{sec:assumptions}

\subsubsection{Teacher error moments (aggregation quality)}

\begin{assumption}[Bias-variance-correlation for weighted aggregation]
\label{ass:teacher}
Fix $x\sim\mathcal D_{\mathrm{pub},x}$ and let $p^*(x)=f(x,\theta^*)$.
Define client prediction errors
\[
\xi_i(x):=q_i(x)-p^*(x),\qquad i\in\mathcal S.
\]
Let $w(x)=(w_i(x))_{i\in\mathcal S}$ be the (possibly data-dependent) aggregation weights with
$w_i(x)\ge 0$ and $\sum_{i\in\mathcal S} w_i(x)=1$.
Define the aggregated error $\bar\xi(x):=\sum_{i\in\mathcal S} w_i(x)\xi_i(x)$.

\paragraph{(Bias term.)}
Define the squared bias functional
\[
B_w^2 := \E_{x}\Big[\big\|\E[\bar\xi(x)\mid x,w(x)]\big\|_2^2\Big].
\]

\paragraph{(Variance and correlation terms.)}
Define centered errors $\tilde\xi_i(x):=\xi_i(x)-\E[\xi_i(x)\mid x,w(x)]$ and assume there exist constants
$\sigma\ge 0$ and $\rho_c\in[0,1]$ such that uniformly in $(x,w(x))$:
\begin{align}
\E\big[\|\tilde\xi_i(x)\|_2^2 \mid x,w(x)\big] &\le \sigma^2, \label{eq:bvc_var}\\
\big|\E\big[\langle \tilde\xi_i(x),\tilde\xi_j(x)\rangle \mid x,w(x)\big]\big|
&\le \rho_c\sigma^2,\qquad i\neq j. \label{eq:bvc_corr}
\end{align}
\end{assumption}

Assumption~\ref{ass:teacher} is a bias-variance-correlation control for the \emph{weighted} teacher error
$\bar\xi(x)=\sum_{i\in\mathcal S} w_i(x)\xi_i(x)$.
The term $B_w^2$ is the (squared) \emph{weighted bias} of the aggregated teacher on public inputs, and is the main place where
a \emph{data-dependent} rule (e.g.\ UWA) can improve over uniform averaging by down-weighting systematically unreliable clients.
The parameters $\sigma^2$ and $\rho_c$ bound conditional second moments and correlations of centered client errors;
$\rho_c=0$ corresponds to conditionally uncorrelated client errors (maximal averaging benefit), while $\rho_c=1$ yields no variance reduction.
Finally, the quantity $\chi(x)=\sum_i w_i(x)^2$ measures weight concentration.

\subsubsection{Stage 2 stochasticity (noise around \texorpdfstring{$\nabla \Jagg$}{nabla Jagg})}

\begin{assumption}[SGD noise for Stage~2]\label{ass:sgd}
There exists $\nu^2\ge 0$ such that for all $t$:
\begin{align}\label{eq:sgd_noise}
\E[g_t\mid \theta_t] &= \nabla \Jagg(\theta_t), \\
\E\big[\|g_t-\nabla\Jagg(\theta_t)\|_2^2 \big|\theta_t\big] &\le \nu^2.
\end{align}
\end{assumption}

Assumption~\ref{ass:sgd} is a standard unbiasedness and bounded variance condition for stochastic gradient estimators used in stochastic approximation / SGD analyses \citep{nemirovski2009robust, ghadimi2016mini}.

\subsubsection{Geometry of \texorpdfstring{$F$}{F}}

\begin{assumption}[Smoothness (and optionally PL later)]\label{ass:smooth_pl}
$F$ is $L_F$-smooth:
\begin{equation}\label{eq:smooth}
\|\nabla F(\theta)-\nabla F(\theta')\|_2 \le L_F\|\theta-\theta'\|_2,\qquad \forall \theta,\theta' \in \RR^d.
\end{equation}
Optionally, for a stronger result, we also assume Polyak-{\L}ojasiewicz (PL) condition \citep{polyak1963gradient, lojasiewicz1963topological}:
\begin{equation}\label{eq:pl}
\|\nabla F(\theta)\|_2^2 \ge 2\mu_F\big(F(\theta)-F^*\big),\qquad \forall \theta\in \RR^d \quad \text{and}\quad  F^*:= F(\theta^*).
\end{equation}
\end{assumption}

Assumption~\ref{ass:smooth_pl} uses $L_F$-smoothness, a standard regularity condition for first-order optimization methods \citep{nesterov2018lectures}.
When we additionally invoke the Polyak-\L ojasiewicz (PL) inequality (second line of Assumption~\ref{ass:smooth_pl}), we follow the classical condition analyzed by \citet{polyak1963gradient}.

\subsubsection{Bridging assumptions between \texorpdfstring{$F$}{F} and public distillation}

Define the \emph{ideal-teacher} objective
\begin{equation}\label{eq:Jstar_def}
\Jstar(\theta):=\E_{x\sim\Dpubx}\Big[\KL(\pstar(x)\|f(x,\theta))\Big].
\end{equation}

\begin{assumption}[Gradient alignment between $F$ and ideal KD]\label{ass:align}
There exist constants $c\in(0,1]$ and $b\ge 0$ such that the following inequality holds for all $t$:
\begin{equation}\label{eq:align_bc}
\E\left[\|\nabla F(\theta_t) - \nabla J_*(\theta_t)\|^2\right] \leq c\E\left[\|\nabla F(\theta_t)\|^2\right] + b^2
\end{equation}
\end{assumption}

\begin{remark}\label{rem:on_grad_align}
By the definition of $p^*(x)$ and $\theta^*$, we have $\nabla J_*(\theta^*) = 0$ and $\nabla F(\theta^*) = 0$. Hence, the above assumption is relatively mild. In practice, as training approaches its final phase, we expect $\theta_t$ to be close to $\theta^*$, implying that $b$ decreases toward the end of Stage~2. For the sake of analytical simplicity, however, we treat the constants $c$ and $b$ as fixed.
\end{remark}

\subsubsection{Score-energy (Jacobian/score control)}

\begin{assumption}[Score-energy bound]\label{ass:score}
There exists a constant $G_{\mathrm{sc}}\ge 0$ such that for all $t$:
\begin{equation}\label{eq:score_energy}
\E\Big[\E_{x\sim\Dpubx}\Big[\sum_{c=1}^C \|\nabla_\theta \log f_c(x,\theta_t)\|_2^2\Big]\Big]\le G_{\mathrm{sc}}^2.
\end{equation}
\end{assumption}

Assumption~\ref{ass:score} upper bounds the Frobenius norm of the \emph{score matrix}
$S(x,\theta)\in\RR^{C\times d}$ with rows $\nabla_\theta\log f_c(x,\theta)^\top$.
In softmax models, $S(x,\theta)$ is controlled by the Jacobian of logits (and feature norms),
so empirically $G_{\mathrm{sc}}^2$ can be supported by measuring average score/Jacobian norms along training. In many softmax models, $\|S(x,\theta)\|_F^2$ can scale like $O(C)$ times a feature-norm term; thus $S^2$ may grow with $C$
unless features/parameters are controlled (e.g., by regularization or by operating in a bounded region).

\subsection{A Hilbert-space identity used in aggregation MSE proofs}

\begin{lemma}[Conditional identity]\label{lem:cond_pythagoras}
Let $Z\in \RR^k$ be square-integrable and $X$ any random variable.
Then
\begin{equation}\label{eq:cond_pythagoras}
\E\|Z\|_2^2 = \E\|\E[Z\mid X]\|_2^2 + \E\|Z-\E[Z\mid X]\|_2^2.
\end{equation}
\end{lemma}

\begin{proof}
Write
\[
Z=\E[Z\mid X] + \big(Z-\E[Z\mid X]\big).
\]
Then
\[
\|Z\|^2
= \|\E[Z\mid X]\|^2 + \|Z-\E[Z\mid X]\|^2
+2\ip{\E[Z\mid X]}{Z-\E[Z\mid X]}.
\]
Taking expectation, the cross term vanishes because
\[
\E\left[\ip{\E[Z\mid X]}{Z-\E[Z\mid X]}\right]
=
\E\left[\E\left[\ip{\E[Z\mid X]}{Z-\E[Z\mid X]}\mid X\right]\right]
=
\E\left[\ip{\E[Z\mid X]}{\E[Z-\E[Z\mid X]\mid X]}\right]=0.
\]
Thus \eqref{eq:cond_pythagoras} holds.
\end{proof}

\subsection{Proof of Lemma~\ref{lem:teacher_mse_weighted} (teacher MSE)}

\begin{proof}
By definition, $p_{\mathrm{agg}}(x)-p^*(x)=\bar\xi(x)$.
Fix $x$ and condition on $(x,w(x))$. Using the conditional variance identity,
\[
\E\big[\|\bar\xi(x)\|_2^2\mid x,w\big]
=
\big\|\E[\bar\xi(x)\mid x,w]\big\|_2^2
+
\E\big[\|\bar\xi(x)-\E[\bar\xi(x)\mid x,w]\|_2^2\mid x,w\big].
\]
By definition of $\tilde\xi_i(x):=\xi_i(x)-\E[\xi_i(x)\mid x,w]$, we have
\[
\bar\xi(x)-\E[\bar\xi(x)\mid x,w]=\sum_{i\in\mathcal S} w_i(x)\tilde\xi_i(x).
\]
Expanding the quadratic form gives
\begin{align*}
\E\Big[\Big\|\sum_{i\in \mathcal S} w_i\tilde\xi_i\Big\|_2^2 \Big| x,w\Big]
&=
\sum_i w_i^2\E[\|\tilde\xi_i\|_2^2\mid x,w]
+
2\sum_{i<j} w_iw_j\E[\langle \tilde\xi_i,\tilde\xi_j\rangle\mid x,w]\\
&\le
\sigma^2\sum_i w_i^2
+
2\rho_c\sigma^2\sum_{i<j} w_iw_j,
\end{align*}
where we used \eqref{eq:bvc_var}-\eqref{eq:bvc_corr}.
Since $\sum_i w_i=1$, we have $2\sum_{i<j}w_iw_j = 1-\sum_i w_i^2 = 1-\chi(x)$, hence
\[
\E\Big[\Big\|\sum_i w_i\tilde\xi_i\Big\|_2^2 \Big| x,w\Big]
\le
\sigma^2\chi(x)+\rho_c\sigma^2(1-\chi(x))
=
\sigma^2\big(\rho_c+(1-\rho_c)\chi(x)\big).
\]
Therefore,
\[
\E\big[\|\bar\xi(x)\|_2^2\mid x,w\big]
\le
\big\|\E[\bar\xi(x)\mid x,w]\big\|_2^2
+
\sigma^2\big(\rho_c+(1-\rho_c)\chi(x)\big).
\]
Taking expectation over $x$ (and any randomness in $w$) yields
\[
\varepsilon_{\mathrm{teach}}
\le
\E_x\big\|\E[\bar\xi(x)\mid x,w(x)]\big\|_2^2
+
\sigma^2\big(\rho_c+(1-\rho_c)\E_x\chi(x)\big)
=
B_w^2+\sigma^2\big(\rho_c+(1-\rho_c)\chi\big),
\]
which is \eqref{eq:teacher_mse_bvc}.
\end{proof}

\subsection{Gradient discrepancy bound (aggregation error \texorpdfstring{$\Rightarrow$}{⇒} gradient error)}

Define the score matrix $S(x,\theta)\in\RR^{C\times d}$ with rows
$[S(x,\theta)]_{c,:}=\nabla_\theta\log f_c(x,\theta)^\top$.

\begin{lemma}[Aggregation-induced gradient discrepancy]\label{lem:agg_grad_app}
Let
\[
\Delta_{\mathrm{agg}}(\theta):=\nabla\Jagg(\theta)-\nabla\Jstar(\theta).
\]
Under Assumption~\ref{ass:score} and for Stage 2 iterates $\theta_t$,
\begin{equation}\label{eq:agg_grad_bound_app}
\E\|\Delta_{\mathrm{agg}}(\theta)\|_2^2
\le
G_{\mathrm{sc}}^2\cdot \E_{x\sim\Dpubx}\|\pagg(x)-\pstar(x)\|_2^2
=
G_{\mathrm{sc}}^2\varepsilon_{\mathrm{teach}}.
\end{equation}
\end{lemma}

\begin{proof}
Fix $\theta$. For any $p(x)\in\Delta^{C-1}$,
\[
\nabla_\theta \KL(p(x)\|f(x,\theta)) = -\sum_{c=1}^C p_c(x)\nabla_\theta \log f_c(x,\theta).
\]
Hence
\begin{align*}
\Delta_{\mathrm{agg}}(\theta)
&=
\nabla\Jagg(\theta)-\nabla\Jstar(\theta) \\
&=
\E_x\left[\sum_{c=1}^C\big({\pstar}_c(x)-{\pagg}_c(x)\big)\nabla_\theta\log f_c(x,\theta)\right] \\
&=
-\E_x\big[S(x,\theta)^\top \bar\xi (x)\big].
\end{align*}

Now bound step-by-step:
\begin{align*}
\|\Delta_{\mathrm{agg}}(\theta)\|
&= \left\|\E_x\big[S(x,\theta)^\top \bar\xi (x)\big]\right\|
\le \E_x\left\|S(x,\theta)^\top \bar\xi (x)\right\| \quad\text{(Jensen's inequality)}\\
&\le \E_x\left[\|S(x,\theta)\|_F\|\bar\xi (x)\|\right] \quad\text{(since }\|A^\top v\|\le\|A\|_F\|v\|\text{)}\\
&\le \sqrt{\E_x\|S(x,\theta)\|_F^2}\cdot \sqrt{\E_x\|\bar\xi (x)\|^2} \quad\text{(Cauchy-Schwarz).}
\end{align*}
Square both sides:
\[
\|\Delta_{\mathrm{agg}}(\theta)\|^2
\le
\left(\E_x\|S(x,\theta)\|_F^2\right)\left(\E_x\|\bar\xi (x)\|^2\right).
\]
Take expectation over Stage~2 randomness (if $\theta$ is random) and use Assumption~\ref{ass:score}:
\[
\E\|\Delta_{\mathrm{agg}}(\theta_t)\|^2
\le
G_{\mathrm{sc}}^2\cdot \E_x\|\bar\xi (x)\|^2
=
G_{\mathrm{sc}}^2\varepsilon_{\mathrm{teach}}.
\]
\end{proof}

\subsection{Proof of Proposition~\ref{prop:implies-ajalloeian}}

\begin{proof}[Proof of Proposition~\ref{prop:implies-ajalloeian}]
We must show the two inequalities in \eqref{eq:bias_bound_main}-\eqref{eq:noise_bound_main}.

\paragraph{Noise.}
By definition $n_t=g_t-\nabla\Jagg(\theta_t)$. Assumption~\ref{ass:sgd} yields
$\E[\|n_t\|^2\mid\theta_t]\le\nu^2$, proving \eqref{eq:noise_bound_main}.

\paragraph{Bias.}
Write
\[
\beta(\theta_t)
=
\nabla\Jagg(\theta_t)-\nabla F(\theta_t)
=
\underbrace{\big(\nabla\Jagg(\theta_t)-\nabla\Jstar(\theta_t)\big)}_{\Delta_{\mathrm{agg}}(\theta_t)}
+
\underbrace{\big(\nabla\Jstar(\theta_t)-\nabla F(\theta_t)\big)}_{e(\theta_t)}.
\]
Use the inequality $\|u+v\|^2\le (1+\alpha)\|u\|^2 + (1+1/\alpha)\|v\|^2$:
\begin{equation}\label{eq:bias_split}
\|\beta(\theta_t)\|^2
\le
(1+\alpha)\|\Delta_{\mathrm{agg}}(\theta_t)\|^2
+\Big(1+\frac{1}{\alpha}\Big)\|e(\theta_t)\|^2.
\end{equation}
Take expectation.

\emph{Step 1: bound $\E\|\Delta_{\mathrm{agg}}(\theta_t)\|^2$.}
By Lemma~\ref{lem:agg_grad_app},
\[
\E\|\Delta_{\mathrm{agg}}(\theta_t)\|^2 \le G_{\mathrm{sc}}^2\varepsilon_{\mathrm{teach}}.
\]

\emph{Step 2: bound $\E\|e(\theta_t)\|^2$ using alignment.}
Take expectation and apply Assumption~\ref{ass:align}:
\[
\E\|e(\theta_t)\|^2 = \E\left[\|\nabla\Jstar(\theta_t)-\nabla F(\theta_t)\|^2\right] \leq c\E\|\nabla F(\theta_t)\|^2 + b^2.
\]

\emph{Step 3: combine.}
Insert the bounds into \eqref{eq:bias_split} and group terms:
\[
\E\|\beta(\theta_t)\|^2
\le
\Big(1+\frac{1}{\alpha}\Big) c \E\|\nabla F(\theta_t)\|^2
+
(1+\alpha)G_{\mathrm{sc}}^2\varepsilon_{\mathrm{teach}}
+\Big(1+\frac{1}{\alpha}\Big)b^2.
\]
This matches \eqref{eq:bias_bound_main} with $m,\zeta^2$ as in \eqref{eq:mz_def_main}.
\end{proof}

\subsection{Convergence rates}\label{sec:convergence_rates_proof}

\begin{theorem}[Nonconvex stationarity under an \emph{expected} biased oracle]
\label{thm:expected_oracle_nonconvex}
Assume $F$ is $L_F$-smooth and Stage~2 updates satisfy
$g_t=\nabla F(\theta_t)+\beta(\theta_t)+n_t$ with $\E[n_t\mid\theta_t]=0$ and
$\E[\|n_t\|^2\mid\theta_t]\le \nu^2$.
Assume there exist $m\in[0,1)$ and $\zeta\ge 0$ such that for all $t$,
\[
\E\|\beta(\theta_t)\|^2 \le m\E\|\nabla F(\theta_t)\|^2+\zeta^2.
\]
If $\gamma\le 1/L_F$ and $\theta_{\mathrm{out}}$ is chosen uniformly at random from $\{\theta_0,\ldots,\theta_{T-1}\}$, then
\begin{equation}\label{eq:expected_oracle_stationarity}
\E\|\nabla F(\theta_{\mathrm{out}})\|^2
\le
\frac{2(F(\theta_0)-F_{\inf})}{T\gamma(1-m)}
+
\frac{L_F\gamma\nu^2}{1-m}
+
\frac{\zeta^2}{1-m},
\end{equation}
where $F_{\inf}:=\inf_\theta F(\theta)$.
\end{theorem}

\begin{proof}
The proof follows the descent template of \citet{ajalloeian2021convergencesgdbiasedgradients} (with expectations moved outside).
By $L_F$-smoothness,
\[
F(\theta_{t+1})\le F(\theta_t)-\gamma\langle\nabla F(\theta_t),g_t\rangle+\frac{L_F\gamma^2}{2}\|g_t\|^2.
\]
Condition on $\theta_t$. Since $\E[n_t\mid\theta_t]=0$,
\[
\E[\langle\nabla F(\theta_t),g_t\rangle\mid\theta_t]
=
\langle\nabla F(\theta_t),\nabla F(\theta_t)+\beta(\theta_t)\rangle.
\]
Also expanding the square and using $\E[n_t\mid\theta_t]=0$ gives
\[
\E[\|g_t\|^2\mid\theta_t]=\|\nabla F(\theta_t)+\beta(\theta_t)\|^2+\E[\|n_t\|^2\mid\theta_t].
\]
Using $\gamma\le 1/L_F$ and combining the inner-product and squared-norm terms yields
\[
\E[F(\theta_{t+1})]
\le
\E[F(\theta_t)]
-\frac{\gamma}{2}\E\|\nabla F(\theta_t)\|^2
+\frac{\gamma}{2}\E\|\beta(\theta_t)\|^2
+\frac{L_F\gamma^2}{2}\nu^2.
\]
Plug $\E\|\beta(\theta_t)\|^2\le m\E\|\nabla F(\theta_t)\|^2+\zeta^2$ and rearrange:
\[
\frac{\gamma}{2}(1-m)\E\|\nabla F(\theta_t)\|^2
\le
\E[F(\theta_t)]-\E[F(\theta_{t+1})]+\frac{\gamma}{2}\zeta^2+\frac{L_F\gamma^2}{2}\nu^2.
\]
Sum over $t=0,\ldots,T-1$ (telescoping) and use $\E[F(\theta_T)]\ge F_{\inf}$. Dividing by $T$ and using the random uniformly distributed output iterate identity yields \eqref{eq:expected_oracle_stationarity}.
\end{proof}

\begin{theorem}[PL linear rate under an \emph{expected} biased oracle]
\label{thm:expected_oracle_pl}
Assume the conditions of Theorem~\ref{thm:expected_oracle_nonconvex} and additionally PL:
$\|\nabla F(\theta)\|^2\ge 2\mu_F(F(\theta)-F^*)$.
Then for any $\gamma\le 1/L_F$,
\begin{equation}\label{eq:expected_oracle_pl}
\E[F(\theta_T)-F^*]
\le
\big(1-\gamma\mu_F(1-m)\big)^T(F(\theta_0)-F^*)
+
\frac{\zeta^2+L_F\gamma\nu^2}{2\mu_F(1-m)}.
\end{equation}
\end{theorem}

\begin{proof}
From the one-step inequality inside the proof of Theorem~\ref{thm:expected_oracle_nonconvex},
\[
\E[F(\theta_{t+1})-F^*]
\le
\E[F(\theta_t)-F^*]-\frac{\gamma}{2}(1-m)\E\|\nabla F(\theta_t)\|^2+\frac{\gamma}{2}\zeta^2+\frac{L_F\gamma^2}{2}\nu^2.
\]
Apply PL: $\E\|\nabla F(\theta_t)\|^2\ge 2\mu_F\E[F(\theta_t)-F^*]$, yielding a linear recursion.
Unrolling it gives \eqref{eq:expected_oracle_pl}.
\end{proof}

\subsection{Iteration complexities}

\begin{proof}[Proof of Corollary~\ref{cor:expected_oracle_bigO_nonconvex}]
Start from the exact nonconvex stationarity bound \eqref{eq:expected_oracle_stationarity}:
\[
\E\|\nabla F(\theta_{\mathrm{out}})\|^2
\le
\frac{2(F(\theta_0)-F_{\inf})}{T\gamma(1-m)}
+
\frac{L_F\gamma\nu^2}{1-m}
+
\frac{\zeta^2}{1-m}.
\]
By definition, $\epsilon_{\min}=\zeta^2/(1-m)$, so it suffices to enforce
\begin{equation}\label{eq:need_bar_eps}
\frac{2(F(\theta_0)-F_{\inf})}{T\gamma(1-m)}
+
\frac{L_F\gamma\nu^2}{1-m}
\le \bar\epsilon.
\end{equation}
We choose $\gamma$ so that the variance term is at most $\bar\epsilon/2$:
\[
\frac{L_F\gamma\nu^2}{1-m} \le \frac{\bar\epsilon}{2}
\quad\Longleftrightarrow\quad
\gamma \le \frac{\bar\epsilon(1-m)}{2L_F\nu^2}.
\]
Together with the stepsize restriction $\gamma\le 1/L_F$ from ~\ref{thm:expected_oracle_nonconvex},
this yields the choice
\[
\gamma := \min\left\{\frac{1}{L_F},\ \frac{\bar\epsilon(1-m)}{2L_F\nu^2}\right\}.
\]
Under this choice, \eqref{eq:need_bar_eps} is implied by requiring the remaining term to be at most $\bar\epsilon/2$:
\[
\frac{2(F(\theta_0)-F_{\inf})}{T\gamma(1-m)} \le \frac{\bar\epsilon}{2}
\quad\Longleftrightarrow\quad
T \ge \frac{4(F(\theta_0)-F_{\inf})}{\gamma(1-m)\bar\epsilon}.
\]
Now substitute the concrete upper bound $\gamma \le \frac{\bar\epsilon(1-m)}{2L_F\nu^2}$ (which holds whenever that branch is active):
\[
T \ge \frac{4(F(\theta_0)-F_{\inf})}{(1-m)\bar\epsilon}\cdot \frac{2L_F\nu^2}{\bar\epsilon(1-m)}
=
\frac{8L_F\nu^2(F(\theta_0)-F_{\inf})}{(1-m)^2\bar\epsilon^2}.
\]
Thus it suffices to take
\[
T = \mathcal O\left(\frac{L_F\nu^2(F(\theta_0)-F_{\inf})}{(1-m)^2\bar\epsilon^2}\right),
\]
which proves the claim.
\end{proof}

\begin{proof}[Proof of Corollary~\ref{cor:expected_oracle_bigO_pl}]
Start from the PL bound \eqref{eq:expected_oracle_pl}:
\[
\E[F(\theta_T)-F^*]
\le
\big(1-\gamma\mu_F(1-m)\big)^T(F(\theta_0)-F^*)
+
\epsilon_{\mathrm{floor}}(\gamma).
\]
Fix a target $\epsilon>\epsilon_{\mathrm{floor}}(\gamma)$ and define $\bar\epsilon:=\epsilon-\epsilon_{\mathrm{floor}}(\gamma)>0$.
It suffices to ensure
\begin{equation}\label{eq:pl_need}
\big(1-\gamma\mu_F(1-m)\big)^T(F(\theta_0)-F^*) \le \bar\epsilon.
\end{equation}
Since $\gamma\le 1/L_F$ and $\mu_F(1-m)>0$, we have $\gamma\mu_F(1-m)\in(0,1)$ for any standard stepsize choice.
Using the inequality $1-x\le e^{-x}$ for $x\in(0,1)$ gives
\[
\big(1-\gamma\mu_F(1-m)\big)^T \le \exp\big(-\gamma\mu_F(1-m)T\big).
\]
Therefore, \eqref{eq:pl_need} is implied by
\[
\exp\big(-\gamma\mu_F(1-m)T\big)(F(\theta_0)-F^*) \le \bar\epsilon,
\]
which is equivalent (taking logarithms) to
\[
T \ge \frac{1}{\gamma\mu_F(1-m)}\log\left(\frac{F(\theta_0)-F^*}{\bar\epsilon}\right)
=
\frac{1}{\gamma\mu_F(1-m)}\log\left(\frac{F(\theta_0)-F^*}{\epsilon-\epsilon_{\mathrm{floor}}(\gamma)}\right).
\]
This yields the stated Big-$\mathcal O$ complexity.
\end{proof}

\paragraph{Code Availability.}
Implementations of all aggregation methods are available at
\url{https://github.com/kovalchuk026/fd_aggregators}.

\end{document}